\documentclass[11pt]{article}
\usepackage{emnlp2021}
\usepackage{amsmath}
\usepackage{times}
\usepackage{gb4e}
\noautomath
\usepackage{pifont}
\newcommand{\cmark}{\ding{51}}%
\newcommand{\xmark}{\ding{55}}%

\usepackage{adjustbox}
\usepackage{booktabs}
\usepackage{dirtytalk}
\usepackage{graphicx}
\usepackage{listings}
\usepackage{microtype}
\usepackage[normalem]{ulem}

\renewcommand{\textsc}[1]{\uppercase{#1}}

\DeclareMathOperator*{\argmax}{\textrm{argmax}}
\newcommand{\example}[1]{\emph{#1}}
\newcommand{\term}[1]{\emph{#1}}
\newcommand{\compose}{\circ}

\title{Structured Abbreviation Expansion in Context}
\author{Kyle Gorman, Christo Kirov, Brian Roark, Richard Sproat \\
Google Inc.}
\date{}

\begin{document}
\maketitle
\begin{abstract}
Ad hoc abbreviations are commonly found in informal communication channels that favor shorter messages.
We consider the task of reversing these abbreviations in context to recover normalized, expanded versions of abbreviated messages.
The problem is related to, but distinct from,
spelling correction,
in that ad hoc abbreviations are intentional and may involve substantial differences from the original words.
Ad hoc abbreviations are productively generated on-the-fly,
so they cannot be resolved solely by dictionary lookup.
We generate a large, open-source data set of ad hoc abbreviations.
This data is used to study abbreviation strategies and to develop two strong baselines for abbreviation expansion.
\end{abstract}

\section{Introduction}
\label{s:introduction}

\term{Text normalization} refers to transformations used to prepare text for downstream processing.
Originally, this term was reserved for transformations mapping between \say{written} and \say{spoken} forms required by technologies like speech recognition and speech synthesis \citep{Sproat:01},
but it is now used for many other transformations,
including normalizing informal text genres found 
on mobile messaging and social media platforms \citep[e.g.,][]{Eisenstein:13,vanderGoot:19}.

Spans of text may require different kinds of normalization depending on their \term{semiotic class} \cite{Taylor:09}
and the requirements of the downstream application.
For example, cardinal numbers such as \example{123} need to be normalized to a spoken form (\example{one hundred twenty three}) for speech processing,
but this is not necessary for many text processing applications.
The class of abbreviations has received particular attention.
High-frequency,
highly-conventionalized abbreviations,
like those used for
units of measure (e.g., \example{mL}, \example{lbs})
or geographic entities (e.g., \example{AK}, \example{NZ})
are often expanded using hand-written grammars
\citep[e.g.,][]{Ebden:Sproat:15},
possibly augmented with machine learning systems for contextual disambiguation 
\citep[e.g.,][]{Ng:17,Zhang:19}.

In this study we are interested in a different subclass of abbreviations,
those which are neither frequent nor conventionalized.
We refer to these as \term{ad hoc abbreviations}.
Such abbreviations are particularly common on those communication channels
which demand or favor brevity,
such as mobile messaging and social media platforms
\citep{Crystal:01,Crystal:08,McCulloch:19}.
Unlike conventionalized abbreviations,
ad hoc abbreviations are an open class,
generated on-the-fly.

Unfortunately, there is little annotated data available to study ad hoc abbreviations as they occur in natural text.
To remedy this, we provide a new open-source data set%
---derived from English Wikipedia---%
designed specifically to collect sentences with ad hoc abbreviations.
We also provide two strong baseline abbreviation expansion systems,
one finite-state, one neural, 
and find that abbreviations in context can be expanded with human-like accuracy.
Both baselines use a noisy channel approach,
which combines an abbreviation model, applied independently to each %out-of-vocabulary
word in the sentence, and a language model enforcing fluency and local coherence
in the expansion.

\begin{figure*}[t]
\centering
\texttt{
\begin{tabular}{llllllllllllll}
the & reason & i & went & to & the & store & was & to & buy & milk & and & bread & . \\
th & rsn & i & went & to & the & str & ws & to & buy & mlk & and & brd & .
\end{tabular}
}
\caption{Example of paired data for this task;
above: expanded sequence; below: abbreviated sequence.
Note that the data has been case-folded.}
\label{t:example}
\end{figure*}

\subsection{Contributions}
\label{ss:contributions}

The contributions of this study are three-fold.
First, we describe a large data set for English abbreviation expansion
made freely available to the research community.
Secondly, we validate this data set using exploratory 
data analysis to identify common abbreviation strategies
used in the training portion of the data set.
Third, we describe and evaluate two strong baseline models%
---one using weighted finite-state transducers,
the other using neural networks---%
and conduct ablation experiments and manual error analyses to study the relative contributions of our various design choices.

\subsection{Related work}
\label{ss:related}

Text normalization was first studied for text-to-speech synthesis (TTS);
\citet{Sproat:01} and \citet{vanEsch:Sproat:17} provide taxonomies of semiotic classes important for speech applications.
Normalization for this application remains a topic of active research \citep[e.g.,][]{Ebden:Sproat:15,Ritchie:19,Zhang:19}. 
\citet{Roark:Sproat:14} focus on abbreviation expansion and enforce a high-precision operating point, since, for TTS, incorrect expansions are judged more costly than leaving novel abbreviations unexpanded.

Abbreviation expansion has also been studied using data from
SMS \citep{Choudhury:07,Beaufort:10},
chatrooms \citep{Aw:Lee:12},
and social media platforms such as Twitter \citep{Chrupala:14,Baldwin:15}.
Most of the prior studies use small, manually curated databases
in which \say{ground truth} labels were generated 
by asking annotators to expand the abbreviations using local context.
\citet{Han:Baldwin:11},
\citet{Yang:Eisenstein:13},
and the organizers of the W-NUT 2015 shared task \citep{Baldwin:15}
all released data sets containing English-language Tweets annotated with expansions for abbreviations.
Unfortunately, none of these data sets are presently available.%
\footnote{
    This may reflect licensing issues inherent to Twitter data.
    This lead us to focus on sources with less restrictive licenses.
}
We are unaware of any large, publicly-available
data set for abbreviation expansion,
excluding a synthetic, automatically-generated data set for informal text normalization \citep{Dekker:vanderGoot:20}.

A wide variety of machine learning techniques
have been applied to abbreviation expansion,
including hidden Markov models,
various taggers and classifiers,
generative and neural language models,
and even machine translation systems.
In this work we focus on supervised models,
though unsupervised approaches have also been proposed
\citep[e.g.,][]{Cook:Stevenson:09,Liu:11,Yang:Eisenstein:13}.
The noisy channel paradigm we use to build baseline models here is inspired by earlier models for contextual spelling correction \citep[e.g.,][]{Brill:Moore:00}.

\subsection{Task definition}
\label{ss:task}

We assume the following task definition.
Let $\mathbf{A} = [a_0, a_1, \ldots, a_n]$ be a sequence of possibly-abbreviated words and let $\mathbf{E}$ be a sequence of expanded words $[e_0, e_1, \ldots, e_n]$, both of length $n$.
If $e_i$ is an element of $\mathbf{E}$,
then the corresponding 
element of $\mathbf{A}$, $a_i$,
must either be identical to $e_i$
(in the case that it is not abbreviated),
or a proper, non-null subsequence of $e_i$ (in the case that it is an abbreviation of $e_i$).
%During training, the overall system is presented with $(\mathbf{A}, \mathbf{E})$
%pairs.
At inference time, the system is presented with an abbreviated $\mathbf{A}$ sequence of length $n$
and is asked to propose a single hypothesis expansion of length $n$, denoted by $\hat{\mathbf{E}}$.

This formulation limits us to abbreviations that are derived via
character deletion,
and forbids pairs such as \example{because} $\rightarrow$ \example{cuz},
which can only be generated via insertion or substitution.
In other words, this task corresponds to what \citet{Pennell:Liu:2010} call \term{deletion-based abbreviation},
arguably the most canonical form of abbreviation in English \citep[e.g.,][]{Cannon:89}.
Furthermore,
by asserting that the abbreviated sentence have the same number of words as the expanded sentence,
we forbid mappings that involve multiple abbreviated or expanded tokens (e.g., \example{to be} $\rightarrow$ \example{2b}).
These restrictions yield a highly-tractable task definition,
but we anticipate that such restrictions can easily be relaxed in future work if desired.

\section{Data}
\label{s:data}

Our goal was to construct a data set consisting of
English abbreviated/expanded sentence pairs as shown in \autoref{t:example}.
In much of the previous work,
these were created by finding text that contains
likely abbreviations, and then asking annotators to disambiguate.
However, \citet{Baldwin:15} reports that this disambiguation task
results in poor inter-annotator agreement.
Therefore, we instead choose to generate data using a task in which
annotators \term{generate} abbreviated text rather than disambiguate it.
Furthermore, since there are many ways to abbreviate any given sentence,
one can easily collect multiple annotations per sentence.

\subsection{Data set construction}
\label{ss:data-set-construction}

We begin with sentences sampled from English-language Wikipedia pages.
We then apply several filters to arrive at sentences in the collection that would fit the abbreviation generation elicitation approach.
As detailed below, annotators are asked to delete at least a minimum number of characters from the sentence while preserving overall intelligibility.
We select sentences of moderate length containing frequent words,
avoiding proper names, technical jargon, and numerical expressions.
Specific filters used to select sentences annotation are:

\begin{itemize}
\item Sentence length $< 150$ characters.
\item Number of words in the sentence $> 8$.
\item Average word length in the sentence $\ge 6$.
\item Sentence must match the regular expression \lstinline|/^[A-Za-z',\-\ ]+\.$/|; i.e., the sentence must consist solely of Latin script letters, whitespace, and a few punctuation marks, including a sentence-final period.
\item All non-initial words must be lowercase tokens (i.e., to avoid sampling proper names).
\end{itemize}

These filters produce a smaller corpus of roughly 4m sentences
and 670k wordtypes.
From this we construct a lexicon of roughly 100k words by retaining all those that occur at least 8 times,
and remove any sentences that contain out-of-vocabulary tokens (OOVs). 
This set of preserved common words still contains many odd words and highly-specialized vocabulary.
We therefore train a byte 5-gram language model from the data and use it to rank sentences by per-character entropy.
%An example sentence excluded from consideration by this filter is: \emph{Tinea versicolor, tinea nigra, white piedra, and black piedra.}
We finally randomly sample 27k sentences with below-median per-character entropy for annotation,
retaining another 2.7m sentences for language model training.

\subsection{Human abbreviation generation}

Ever since the earliest written texts,
scribes have used ad hoc abbreviations to minimize space and time.
Indeed, anyone who has studied ancient inscriptions is struck by the extremely high rate of abbreviation in such texts.
For example, 11 of the 35 tokens in the dedicatory inscription at the base of Trajan's Column, completed in 113 CE, are%
---largely ad hoc---%
abbreviations (\autoref{f:trajan}).

\begin{figure}
\begin{adjustbox}{max width=\columnwidth}
\begin{tabular}{l}
\textsc{senatvs popvlvsqve romanvs} \\ % 3
\textsc{\uline{imp} caesari divi nervae \uline{f} nervae} \\ % 6
\textsc{traiano \uline{avg} \uline{germ} dacico \uline{pontif}} \\ % 5
\textsc{maximo \uline{trib} \uline{pot} xvii \uline{imp} vi \uline{cos} vi \uline{p} \uline{p}} \\ % 10
\textsc{ad declarandvm qvantae altitvdinis} \\ % 4
\textsc{mons et locvs tan}[tis oper]\textsc{ibvs sit egestvs} \\ % 7
\end{tabular}
\end{adjustbox}
\caption{
Latin dedicatory inscription at the base of Trajan's Column
(\emph{CIL} 6.960; \citealp{CIL6}),
commemorating the Roman victory in the Dacian Wars,
with abbreviations underlined.
}
\label{f:trajan}
\end{figure}

With this in mind, we designed an annotation task in which
a team of six in-house professional annotators, each working independently,
were instructed to remove at least 20 characters from each sentence while maintaining overall intelligibility.
A custom browser-based annotation interface is used to enforce the task limitations described in \autoref{ss:task},
namely that abbreviations can only be produced by deletion
and that no token can be totally deleted.
It was expected that this annotation procedure would produce high rates of ad hoc abbreviation use%
---higher than is likely to occur naturally---%
and that subsequent expansion could be made less challenging by replacing a random fraction of the abbreviated tokens with their corresponding expansions,
creating a corpus with a lower rate of abbreviation and reducing overall ambiguity.
Annotators are provided no information about the intended use of this corpus.

The abbreviated/expanded sentence pairs are then randomly
partitioned into training (80\%), development (10\%), and testing (10\%) sets.
Summary statistics for the data set are given in \autoref{t:summary}.
Note that some sentences in the training set are deliberately abbreviated by multiple annotators;
these are considered separate sentences for the purposes of this table.

\begin{table}
\centering
\begin{tabular}{l c rr}
\toprule
            && \# sentences & \# tokens \\
\cmidrule{3-4}
Training    &&       21,318 &   332,829 \\
Development &&        2,665 &    41,757 \\
Testing     &&        2,665 &    41,730 \\
\cmidrule{3-4}
LM data     &&    2,657,826 & 41,573,540 \\
\bottomrule
\end{tabular}
\caption{Summary statistics for the data set.}
\label{t:summary}
\end{table}

\subsection{Exploratory analysis}

\begin{table}[t]
\centering
\begin{tabular}{rrr}
\toprule
deletions & count   & \% \\
\midrule
        0 & 182,552 & 54.8 \\
        1 & 78,872  & 23.7 \\
        2 & 42,976  & 12.9 \\
        3 & 17,105  &  5.1 \\
  $\ge$ 4 & 11,324  &  3.4 \\
\bottomrule
\end{tabular}
\caption{Histogram giving the number of deletions per token in the training set.}
\label{tab:histogram}
\end{table}

\begin{table*}[t]
\centering
\begin{tabular}{l lcl r}
\toprule
strategy & \multicolumn{3}{l}{example} & \% \\
\midrule
delete final \emph{e}      & {native}  & $\rightarrow$ & {nativ}        & 12.0 \\
delete other final letter  & {jamming} & $\rightarrow$ & {jammin}       &  2.3 \\
delete 2 final letters     & {however} & $\rightarrow$ & {howev}        &  0.6 \\
delete 3 final letters     & {volume}  & $\rightarrow$ & {vol}          &  1.2 \\
delete 4 final letters     & {develop} & $\rightarrow$ & {dev}          &  1.6 \\
\multicolumn{4}{l}{(total)}                                             & 17.6 \\
\\
delete all vowels           & {government}& $\rightarrow$ & {gvrnmnt}   & 26.2 \\
delete all but word-initial & {unheard}   & $\rightarrow$ & {unhrd}     & 10.9 \\
delete all but first vowel  & {municipal} & $\rightarrow$ & {muncpl}    &  9.3 \\
delete all but final vowel  & {testing}   & $\rightarrow$ & {tsting}    &  3.8 \\
delete other vowel subsets  & {reviewers} & $\rightarrow$ & {rviewrs}   & 18.1 \\
\multicolumn{4}{l}{(total)}                                             & 68.3 \\
\\
delete all vowels \& other       & {background} & $\rightarrow$ & {bkgrnd}   &  3.7 \\
delete duplicated consonants     & {accessible} & $\rightarrow$ & {acesible} &  2.0 \\
delete non-duplicated consonants & {meetings}   & $\rightarrow$ & {meetins}  &  1.2 \\
other                            & {often}      & $\rightarrow$ & {ofn}      &  7.3 \\
\multicolumn{4}{l}{(total)}                                             & 14.2 \\
\bottomrule
\end{tabular}
\caption{
Percentages of the 150k training set abbreviations following three major abbreviation strategies:
suffix deletion, vowel deletion, and other strategies.
}
\label{tab:strategies}
\end{table*}

To validate this novel annotation process we conducted an exploratory analysis focusing on common abbreviation patterns used by the annotators.
As shown in \autoref{tab:histogram},
over 45\% of the training set tokens are abbreviated,
and just over half of these have just one character deleted.
This suggests that annotators frequently choose to make small changes to many words
rather than making more aggressive abbreviations of a smaller number of words.
Beyond single-character deletions,
the most common strategies involve the deletion of a (string) suffix or (orthographic) vowels,
as shown in \autoref{tab:strategies}.
Simply eliding all orthographic vowels%
---and preserving all consonants---in a word is the single most common specific strategy.
This strategy accounts for over a quarter of all training set abbreviations.
Over 80\% of the abbreviations found in the training set preserve all consonants.
These results broadly accord with our intuitions about abbreviation formation in English.

\subsection{Human abbreviation expansion}

To further validate the annotation process
and to establish a human topline,
a separate team of three in-house annotators, each working independently,
attempted to expand abbreviated sentences from the test set.
As was the case for abbreviation generation,
task restrictions were enforced using a custom browser-based annotation interface.
These results are presented below in \autoref{s:results},
but anticipating the findings there,
the second group of annotators were able to recover the original sentence with a high degree of accuracy.

\subsection{Release}

We release all annotated data under the Creative Commons Attribution-ShareAlike 3.0 Unported (CC BY-SA) License, the same license used by Wikipedia itself.%
\footnote{
    \url{https://github.com/google-research-datasets/WikipediaAbbreviationData}
}
The release includes training, development, and testing data in the form of text-format Protocol Buffers messages,%
\footnote{
    \url{https://developers.google.com/protocol-buffers}
}
as well as instructions for deserializing these messages.

\section{Generative story}

We approach the problem of abbreviation expansion as an instance of the noisy channel problem
that has been applied to a wide range of problems
including speech recognition \citep{Jelinek:97,Mohri:02}
and spelling correction \citep{Brill:Moore:00}.
We first describe the generative process that produces 
the abbreviated sentence:

\begin{enumerate}
\item First, generate the expanded sentence $\textbf{E} = 
[e_0, e_1, \ldots, e_n]$
\item Then, generate $\textbf{A} = [a_0, a_1, \ldots, a_n ]$ such that each element $a_i$ is either
    \begin{enumerate}
    \item a non-null proper subsequence of $e_i$ 
    (i.e., $a_i$ abbreviates $e_i$), or
    \item equivalent to $e_i$ (i.e., $a_i = e_i$).
    \end{enumerate}
\end{enumerate}

Given $\textbf{A}$,
which we assume has passed through this noisy channel,
our goal is to recover $\textbf{E}$.
We can naturally express this as a conditional model
using Bayes' theorem.

\begin{align}
\hat{\textbf{E}} &= \argmax_\textbf{E} P(\textbf{E} \mid \textbf{A}) \\
                 &= \argmax_\textbf{E} P(\textbf{E}) \cdot P(\textbf{A} \mid \textbf{E})
\end{align}

\noindent
$P(\textbf{E})$,
the probability of the expanded sequence,
is naturally expressed by a language model over such sequences.
For $P(\textbf{A} \mid \textbf{E})$,
we make the simplifying assumption that the abbreviation of each token%
---or indeed, whether it is abbreviated at all---%
is independent of all other tokens.
This allows us to approximate $P(\textbf{E} \mid \textbf{A})$ as the product

\begin{equation}
P(\textbf{A} \mid \textbf{E}) = \prod_{i = 0}^n P(a_i \mid e_i) .
\label{e:conditional}
\end{equation}
 
\noindent
Under these assumptions, a model for abbreviation expansion is parameterized by an expansion language model $P(\textbf{E})$
and a per-token abbreviation generation model $P(a \mid e)$.

Below we describe methods for constructing language models and abbreviation models and show how these are used to 
infer the expanded sentence for a sentence containing abbreviations.

\section{Models}
\label{s:models}

We propose two baseline systems for noisy-channel decoding,
one that relies on weighted finite-state transducers
and one that uses a neural network language model.

\subsection{Finite-state pipeline}
\label{ss:finite-state-pipeline}

The finite-state pipeline is defined by two weighted finite state automata.
The first is a conventional n-gram language model
defining a probability distribution over expansions

\begin{equation}
P(\textbf{E}) = \prod_{i = 0}^n P(e_i \mid h_i)
\end{equation}

\noindent
where $h_i$, the expansion history, is a finite suffix of $e_0, \ldots, e_{i - 1}$.
The second term is represented by a 
type of weighted finite-state transducer known variously as a
\term{joint multigram model}, 
\term{pair n-gram model}, or
\term{pair language model} (pair LM).
Such models have been used for grapheme-to-phoneme conversion \citep{Bisani:Ney:08,Novak:16},
transliteration \citep{Hellsten:17,Merhav:Ash:18},
and abbreviation expansion \citep{Roark:Sproat:14}
among other tasks.

A pair LM $\alpha$ is a joint model over input/output strings $P(a_i, e_i)$ 
where $a_i$ is an abbreviation and $e_i$ an expansion.
To train the pair LM, one first uses expectation maximization or related algorithms to align
the characters of an abbreviation to its expansion.
For example, for the pair \emph{brd} $\rightarrow$ \emph{bread},
the alignment might be
$[\texttt{b:b},
\texttt{r:r},
\epsilon\texttt{:e},
\epsilon\texttt{:a},
\texttt{d:d}]$
where $\epsilon$ stands in for the empty string.
Then, these alignments are used to construct a conventional n-gram language model
representing the joint probability over input/output pairs (e.g., $\texttt{b}:\texttt{b}$).%
\footnote{
    Computing the conditional probability $P(a_i \mid e_i)$ in eq.~\ref{e:conditional}
    from the joint probability $P(a_i, e_i)$
    requires a computationally expensive summation over all possible alignments.
    However we find it can be effectively approximated using the most probable alignment according to the joint probability model.
}

This model is applied to an abbreviated sentence as follows.%
\footnote{
    We assume the reader is familiar with finite-state automata and algorithms such as
    composition, concatenation, projection, 
    and shortest path.
    See \citealt{Mohri:09} for a review of finite-state automata and these algorithms.
}
First, the abbreviated sentence $\textbf{A}$ is encoded as an unweighted acceptor,
composed with the closure of the pair LM $\alpha$,
and the result is output-projected
(here indicated by $\pi_o$).

\begin{equation}
\eta = \pi_0[\textbf{A} \compose \alpha^*] .
\end{equation}

\begin{figure*}
\centering
\includegraphics[width=\textwidth]{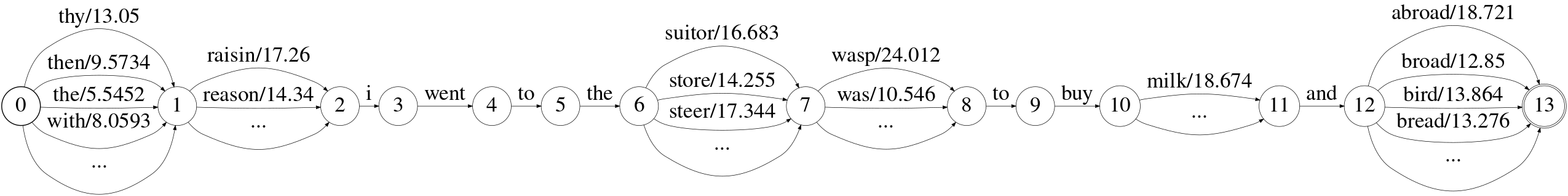}
\caption{
An example $\eta$ lattice corresponding to the example sentence in \autoref{t:example}.
Ellipses indicate that arcs have been pruned for reasons of space.
Non-zero costs are indicated by negative log
probability arc weights.
}
\label{t:sausage}
\end{figure*}

\noindent
An example of the resulting lattice is shown in \autoref{t:sausage}.
$\lambda$ is an unweighted transducer in which each path maps an in-vocabulary word,
encoded as a character sequence, to that same word encoded as a single symbol,
as is done in the expansion LM.
To construct the final lattice,
$\eta$ is composed with the closure of $\lambda$,
output-projected, and composed with the expansion language model $\mu$.
The best expansion is given by

\begin{equation}
\hat{\textbf{E}} = \textrm{ShortestPath}[\pi_o[\eta \compose \lambda^*] \compose \mu],
\end{equation}

\noindent
the shortest path through a weighted lattice of candidate expansions.

\subsection{Neural pipeline}
\label{ss:neural-pipeline}

The neural pipeline replaces the n-gram language model with a recurrent neural network language model.
Unlike conventional n-gram models,
recurrent neural models do not in principle impose an upper bound on the amount of context used to condition their predictions.
While the pair LM can be fused with a neural LM, 
an alternative abbreviation model was also considered.
As before, we wish to generate a set of expansion candidates and allow the language model to select the best candidate in context.
Any in-vocabulary word is considered to be a candidate for $e_i$ so long as it is a supersequence of $a_i$.
A list of such candidates can be generated by constructing
a transducer $\nu$ that allows for identity or insertion relations.
Then

\begin{equation}
\pi_o[a_i \compose \nu \compose \lambda]
\end{equation}

\noindent
contains all possible expansions of $a_i$.

We assign weights to the operations of $\nu$.
Identity mappings are given zero cost,
whereas insertion costs are 
given by the negative log probability of 
that character's insertion, estimated
using maximum likelihood estimation on the training set.
Probabilities for initial and final insertions are computed separately from word-internal insertions.
These weights allow the system to rank candidate expansions at each position.
Only the 8 best candidates
are considered at each position,
and candidates whose path weights are more than twice the cost of the best candidate path are pruned.
A few additional heuristics are used to represent the tension between brevity and fidelity.

\begin{itemize}
\item \textbf{LexBlock:}
If $a_i$ is in-vocabulary,
set the probability of all other output candidates to zero.
\item \textbf{Memory:}
Do not prune an expansion candidate $e_i$ if it is occurs as an expansion of $a_i$ in the training set.
\item \textbf{SubBlock:}
If one candidate is a contiguous substring of another,
set the probability of the superstring candidate to zero.
For example, for the abbreviation \example{ct},
this heuristic will discard the candidate \example{cats} in favor of \example{cat}.
\end{itemize}

\noindent
We refer to this as the \term{subsequence model} to contrast it with the pair LM proposed earlier.

Decoding of the neural pipeline is similar to that of the finite-state pipeline
with the addition of pruning and the optional application of the above heuristics.
However, finding the highest-probability path according to the
neural language model is somewhat more challenging than is the case for the finite-state model.
Because there is no upper bound on the amount of context used by the neural language model,
the score for each node of the lattice depends on the full path taken to reach that node, and the decoding graph is a prefix tree of all paths through the lattice.
As the number of such paths grows exponentially as a function of sentence length, left-to-right approximate beam search \citep{Graves:12} 
with a beam of size 20 is used as an alternative to exhaustive search.

\section{Experiments}
\label{s:experiments}

We perform experiments with both pipelines described above,
the finite state and neural pipelines described in \autoref{s:models}, and compare their performance with human participants attempting to expand the same abbreviated text.
The training set described in \autoref{s:data} is used to train the abbreviation models.
Language models are trained using the concatenation of the training set with 2.7m additional sentences from Wikipedia as described in \autoref{ss:data-set-construction}.
The development set was used to tune the Markov order of the finite-state components,
and to ablate the subsequence model heuristics.
Final evaluations are conducted on the test set.
The full vocabulary consists of all 75k wordtypes appearing in the language model training set,
simulating a general-domain normalization task.

\subsection{Finite-state implementation}
\label{ss:finite-state}

\paragraph{Expansion model} 
The expansion model is a conventional language model over expansion tokens.
The OpenGrm-NGram toolkit \citep{Roark:12}
is used to build a trigram model
with Kneser-Ney smoothing \citep{Ney:94}
and $\epsilon$-arcs used to approximate back-offs.
It is then shrunk using relative entropy pruning \citep{Stolcke:98}.

\paragraph{Abbreviation model}
The abbreviation model is a pair n-gram language model over input/output character pairs,
encoded as a weighted transducer.
The OpenGrm-BaumWelch toolkit and a stepwise interpolated variant of expectation maximization \citep{Liang:Klein:09}
are used to compute alignments between abbreviations and expansions.
OpenGrm-NGram is then used to train a 4-gram pair LM.
As with the expansion model, Kneser-Ney smoothing with $\epsilon$-arc back-offs are used, but no shrinking is performed.%
\footnote{
    We conducted several other experiments that had negative or negligible results,
    including the use of other smoothing techniques,
    using $\phi$-arcs to exactly encode language model back-offs,
    and a finite-state implementation of the subsequence model's LexBlock heuristic.
}

\subsection{Neural implementation}
\label{ss:neural}

\paragraph{Expansion model}
The expansion model consists of an embedding layer of dimensionality 512 and two LSTM \citep{Hochreiter:Schmidhuber:97} layers, each with 512 hidden units.
Each sentence is padded with reserved start and end symbols.
The model, implemented in TensorFlow \citep{Abadi:16},
is trained in batches of 256 until convergence using the Adam optimizer \citep{Kingma:Ba:15} with $\alpha = .001$.

\paragraph{Abbreviation model}
The subsequence model parameters are computed over the training set via maximum likelihood estimation.

\subsection{Complexity}

The complexity of both pipelines is dominated by their decoding step.
The finite-state pipeline's shortest-path computation has complexity of
$O(n \log n)$, where $n$ is the length of the sequence to be decoded.
Beam search for the neural network pipeline has complexity of $O(n)$.

\subsection{Evaluation} 
\label{ss:evaluation}

The primary metric used for system comparison is word error rate (WER),
the percentage of incorrect words in the expansion.
We also compute more specific statistics:
overexpansion rate (OER),
the percentage of words in the hypothesis expansion
which were expanded but did not require expansion,
underexpansion rate (UER),
the percentage of words 
which required expansion but were not expanded,
and incorrect expansion rate (IER),
the percentage of words which both required and received expansion
but which were were expanded incorrectly.%
\footnote{
    Note that UER and IER are calculated using the total words that \emph{should} be expanded as a denominator,
    whereas OER is calculated using the total number of words that should \emph{not} be expanded as a denominator.
    As a result, WER is not merely the sum of OER, IER, and UER.
}
As \citet{Roark:Sproat:14} argue,
an ideal abbreviation expansion system should be \say{Hippocratic} in the sense that it does no harm to human interpretability,
so it is particularly important to minimize OER and IER errors.
Other metrics such as character-level edit distance
and sentence error rate are also computed.
However, they are closely correlated with WER and are therefore omitted below.

\section{Results}
\label{s:results}

\subsection{Test results}

\begin{table*}
\centering
\begin{tabular}{l c rrrr}
\toprule
                   && WER           & OER  & UER  & IER  \\
\midrule
n-gram LM, pair LM && 2.90          & 0.00 & 2.13 & 4.08 \\ 
% LSTM LM, pair LM   && 1.35          & 0.43 & 0.25 & 2.12 \\ %NOTE: results on dev, pairlm order 3 :DEV RESULTS
LSTM LM, pair LM   && 1.41          & 0.39 & 0.19 & 2.35 \\ %NOTE: results on dev, pairlm order 3 :TEST RESULTS
%LSTM LM, subseq.   && \textbf{1.07} & 0.43 & 0.23 & 1.55 \\ % DEV RESULTS
LSTM LM, subseq.   && \textbf{1.12} & 0.40 & 0.20 & 1.74 \\ % TEST RESULTS
Human topline      && 3.51          & 2.23 & 0.30 & 4.88 \\
\bottomrule
%LSTM LM, subseq., task vocab.   && 3.85          & 3.08          & \textbf{0.14} & 4.57 \\
%LSTM LM, pair LM, task vocab.   && 4.09          & 3.10          & 0.15          & 4.57 \\ %NOTE: results on dev, pairlm order 3
\\
\end{tabular}
\caption{
Baseline results, with a human topline for comparison.
WER: word error rate;
OER: overexpansion rate;
UER: underexpansion rate;
IER: incorrect expansion rate.
}
\label{t:results}
\end{table*}

\autoref{t:results} gives an overview of results across the different experimental conditions
as well as the human topline results.
The best overall performance is achieved by the neural pipeline combined with the subsequence abbreviation model.
Presumably the neural pipeline benefits from the more expressive model of local context, and the subsequence model outperforms the pair LM.

%\begin{table*}
%\centering
%\begin{tabular}{l c rrrr}
%\toprule
%                  &&        & +LexBlock & +Memory & +SubBlock \\
%\midrule
%full vocab.       &&   7.92 &      9.16 &    1.18 &     1.07 \\
%task vocab.       &&   9.56 &      8.47 &    3.94 &     3.85 \\
%\bottomrule
%\end{tabular}
%\caption{
%Ablation experiments for subsequence model heuristics, with results in WER.
%}
%\label{t:ablation}
%\end{table*}

\begin{table}
\centering
\begin{tabular}{l c rr}
\toprule
                 && full vocab. & task vocab. \\
\cmidrule{3-4}
Subsequence      &&        7.92 &        9.56 \\
\ldots +LexBlock &&        9.16 &        8.47 \\
\ldots +Memory   &&        1.18 &        3.94 \\
\ldots +SubBlock &&        1.07 &        3.85 \\
\bottomrule
\end{tabular}
\caption{
Development set WER results for the ablation experiments for the subsequence model heuristics.
}
\label{t:ablation}
\end{table}

\subsection{Ablation results}

To measure the importance of the three heuristics used in the subsequence model, a series of ablation experiments are performed on the development set;
results are given in \autoref{t:ablation}.
These show that the best performance was achieved after all heuristics discussed in \autoref{ss:neural-pipeline} are applied to the subsequence model.
The ablation experiments are repeated with a smaller
\say{task vocabulary} containing the 15k wordtypes occuring in the abbreviation training corpus.
One interesting phenomenon is the apparent rise in error rate when the LexBlock heuristic is used with a full vocabulary.
This is primarily due to the fact that the full vocabulary already includes abbreviated function words.
Without also applying the Memory heuristic,
LexBlock requires these common abbreviations to remain unexpanded.
However, this is no longer an issue when the task vocabulary is used in place of the full vocabulary.
 
\subsection{Error analysis}
\label{ss:error-analysis}

Using the development set, we perform a qualitative analysis focusing on overexpansion and incorrect expansion errors made by the best-performing system,
the neural LM with a subsequence abbreviation model.
The examples below give the
\textbf{source abbreviation},
the \textbf{\cmark~target expansion},
and the \textbf{\xmark~predicted expansion}.
They also give the reader an idea of the difficulty of the abbreviation expansion task in the presence of a high rate of abbreviation.
A manual inspection of the 400 errors produced suggests that roughly 40\% could be classified as harmful in the sense that they substantially modify the meaning of the underlying sentence.

\begin{exe}
\ex the \textbf{\{clases, \cmark~classes, \xmark~clashes\}} cntinud nd th band strugld fr time to rite tgthr .
\ex anothr criticism is abt th absenc o a stndrd \textbf{\{auditin, \cmark~auditing, \xmark~audition\}} procedr .
%\ex origly , oonly one chapt ws contemplated by the \textbf{\{fndrs, \cmark founders, \xmark funders\}} \\
\end{exe}

Sometimes these errors are difficult to avoid given a highly ambiguous context for a short abbreviation, with multiple plausible expansions.
A broader, multi-sentence context might help further disambiguate these cases but would naturally require a more complex language model.
Furthermore, 39.8\% of \emph{all} errors are unavoidable
due the aggressive candidate pruning in the best performing conditions.
The expected candidate is not an option for the model in these cases,
though the ablation results suggest this is a sensible trade-off to make.
The remaining errors are largely benign
and showed several re-occurring patterns.
One common problem is the model incorrectly choosing 
an unexpected American and or British spelling variant%
---both are present on Wikipedia---%
an easily-fixed inconsistency in the data.

\begin{exe}
\ex consequently th village hs develpd a mor suburbn role than som o its \textbf{\{neighbrs, \cmark~neighbours, \xmark~neighbors\}} .
%\ex thes seprat negotitns wth th othr emplyrs left the dock \textbf{\{labrers, \cmark labourers, \xmark laborers\}} \\
\end{exe}

\noindent
It is also common for short abbreviations to be
incorrectly expanded to a morphologically-related variant of 
expected expansion,\footnote{
    We note that \citet{Zelasko:18} considers the problem of disambiguating abbreviations in Polish, a language with far richer inflectional morphology.
}
or to a function word with comparable syntactic effect.

\begin{exe}
\ex they \textbf{\{recog, \cmark~recognized, \xmark~recognize\}} accomps by musicians frm th prev yr .
\ex \textbf{\{th, \cmark~the, \xmark~this\}} behavr s strengthnd by an automatc reinfrcng consequenc .
%\ex it is also a market city for the srrdng \textbf{\{agric, \cmark agricultural, \xmark agriculture\}} industry 
%\ex he advocated a \textbf{\{meth, \cmark methodogy, \xmark method\}} tht required validation ginst real vis phenomena 
%\ex no ofical artwrk hs ben presntd fr \textbf{\{th, \cmark the, \xmark this\}} singl
\end{exe}

\section{Ethical concerns}

The proposed technology is intended as a component of other speech and language processing systems.
We note that abbreviation expansion systems have some small potential for abuse
beyond those of the larger systems they might be integrated into.
For instance, this technology could be used to defeat abbreviation as a strategy for circumventing algorithmic state censorship.

The data is drawn from English Wikipedia text and was produced by a team of professional annotators based in the United States;
its use to disambiguate abbreviations generated by other English-speaking communities would likely introduce bias.

\section{Conclusions}

We introduce a large, freely-available data set for ad hoc abbreviation expansion, describing the validating the annotation paradigm used to develop it.
Using this data set,
we find that ad hoc abbreviation expansion can be performed at
human levels of accuracy using noisy channel models.
The finite-state pipeline described above has been integrated as an optional module for Google text-to-speech synthesis engines.

In future work we will survey abbreviation and abbreviation expansion beyond English.
It is expected that abbreviation strategies may differ substantially across languages and scripts.
Indeed, while they are integral features of some languages,
particularly in informal genres,
others appear to use few if any abbreviations at all.

\section*{Acknowledgments}

The authors thank Olivia Redfield for assistance with data collection,
and Caterina Golner and Katherine Wang for their help with data cleaning and pilot experiments.

\bibliography{abbrev}

\providecommand{\noopsort}[1]{}
\begin{thebibliography}{43}
\expandafter\ifx\csname natexlab\endcsname\relax\def\natexlab#1{#1}\fi

\bibitem[{Abadi et~al.(2016)Abadi, Barham, Chen, Chen, Davis, Dean, \ldots, and
  Zheng}]{Abadi:16}
Martín Abadi, Paul Barham, Jianmin Chen, Zhifeng Chen, Andy Davis, Jeffrey
  Dean, \ldots, and Xiaoqiang Zheng. 2016.
\newblock {TensorFlow}: a system for large-scale machine learning.
\newblock In \emph{12th {USENIX} Symposium on Operating Systems Design and
  Implementation}, pages 265--283.

\bibitem[{Aw and Lee(2012)}]{Aw:Lee:12}
Ai~Ti Aw and Lian~Hau Lee. 2012.
\newblock Personalized normalization for a multilingual chat system.
\newblock In \emph{Proceedings of the {ACL} 2012 System Demonstrations}, pages
  31--36.

\bibitem[{Baldwin et~al.(2015)Baldwin, Kim, de~Marneffe, Ritter, Han, and
  Xu}]{Baldwin:15}
Timothy Baldwin, Young-Bum Kim, Marie~Catherine de~Marneffe, Alan Ritter,
  Bo~Han, and Wei Xu. 2015.
\newblock Shared tasks of the 2015 workshop on noisy user-generated text:
  {T}witter lexical normalization and named entity recognition.
\newblock In \emph{Proceedings of the Workshop on Noisy User-generated Text},
  pages 126--136.

\bibitem[{Beaufort et~al.(2010)Beaufort, Roekhaut, Cougnon, and
  Fairon}]{Beaufort:10}
Richard Beaufort, Sophie Roekhaut, Louise-Amélie Cougnon, and Cédrick Fairon.
  2010.
\newblock A hybrid rule/model-based finite-state framework for normalizing
  {SMS} messages.
\newblock In \emph{Proceedings of the 48th Annual Meeting of the Association
  for Computational Linguistics}, pages 770--779.

\bibitem[{Bisani and Ney(2008)}]{Bisani:Ney:08}
Maximilian Bisani and Hermann Ney. 2008.
\newblock Joint-sequence models for grapheme-to-phoneme conversion.
\newblock \emph{Speech Communication}, 50(5):434--451.

\bibitem[{Brill and Moore(2000)}]{Brill:Moore:00}
Eric Brill and Robert~C. Moore. 2000.
\newblock An improved error model for noisy channel spelling correction.
\newblock In \emph{Proceedings of the 38th Annual Meeting of the Association
  for Computational Linguistics}, pages 286--293.

\bibitem[{Cannon(1989)}]{Cannon:89}
Garland Cannon. 1989.
\newblock Abbreviations and acronyms in {English} word-formation.
\newblock \emph{American Speech}, 64(2):99--127.

\bibitem[{Choudhury et~al.(2007)Choudhury, Saraf, Jain, Sarkar, and
  Basu}]{Choudhury:07}
Monojit Choudhury, Rahul Saraf, Vijit Jain, Sudesha Sarkar, and Anupam Basu.
  2007.
\newblock Investigation and modeling of the structure of texting language.
\newblock \emph{International Journal of Document Analysis and Recognition},
  10:157--174.

\bibitem[{Chrupała(2014)}]{Chrupala:14}
Grzegorz Chrupała. 2014.
\newblock Normalizing tweets with edit scripts and recurrent neural embeddings.
\newblock In \emph{Proceedings of the 52nd Annual Meeting of the Association
  for Computational Linguistics (Volume 2: Short Papers)}, pages 680--686.

\bibitem[{Cook and Stevenson(2009)}]{Cook:Stevenson:09}
Paul Cook and Suzanne Stevenson. 2009.
\newblock An unsupervised model for text message normalization.
\newblock In \emph{Proceedings of the Workshop on Computational Approaches to
  Linguistic Creativity}, pages 71--78.

\bibitem[{Crystal(2001)}]{Crystal:01}
David Crystal. 2001.
\newblock \emph{Language and the Internet}.
\newblock Cambridge University Press.

\bibitem[{Crystal(2008)}]{Crystal:08}
David Crystal. 2008.
\newblock \emph{Txtng: The Gr8 Db8}.
\newblock Oxford University Press.

\bibitem[{Dekker and {\noopsort{Goot}{van der
  Goot}}(2020)}]{Dekker:vanderGoot:20}
Kelly Dekker and Rob {\noopsort{Goot}{van der Goot}}. 2020.
\newblock Synthetic data for {English} lexical normalization: how close can we
  get to manually annotated data?
\newblock In \emph{Proceedings of the 12th Language Resources and Evaluation
  Conference}, pages 6300--6309.

\bibitem[{Ebden and Sproat(2015)}]{Ebden:Sproat:15}
Peter Ebden and Richard Sproat. 2015.
\newblock The {Kestrel} {TTS} text normalization system.
\newblock \emph{Natural Language Engineering}, 21(3):333--353.

\bibitem[{Eisenstein(2013)}]{Eisenstein:13}
Jacob Eisenstein. 2013.
\newblock What to do about bad language on the internet.
\newblock In \emph{Proceedings of the 2013 Conference of the North American
  Chapter of the Association for Computational Linguistics: Human Language
  Technologies}, pages 359--369.

\bibitem[{{\noopsort{Esch}{van Esch}} and Sproat(2017)}]{vanEsch:Sproat:17}
Daan {\noopsort{Esch}{van Esch}} and Richard Sproat. 2017.
\newblock An expanded taxonomy of semiotic classes for text normalization.
\newblock In \emph{Proceedings of INTERSPEECH}, pages 4016--4020.

\bibitem[{{\noopsort{Goot}{van der Goot}}(2019)}]{vanderGoot:19}
Rob {\noopsort{Goot}{van der Goot}}. 2019.
\newblock {MoNoise}: a multi-lingual and easy-to-use lexical normalization
  tool.
\newblock In \emph{Proceedings of the 57th Annual Meeting of the Association
  for Computational Linguistics: System Demonstrations}, pages 201--206.

\bibitem[{Graves(2012)}]{Graves:12}
Alex Graves. 2012.
\newblock Sequence transduction with recurrent neural networks.
\newblock Paper presented at the Representation Learning Workshop, ICML 2012.

\bibitem[{Han and Baldwin(2011)}]{Han:Baldwin:11}
Bo~Han and Timothy Baldwin. 2011.
\newblock Lexical normalisation of short text messages: makn sense a \#twitter.
\newblock In \emph{Proceedings of the 49th Annual Meeting of the Association
  for Computational Linguistics: Human Language Technologies}, pages 368--378.

\bibitem[{Hellsten et~al.(2017)Hellsten, Roark, Goyal, Allauzen, Beaufays,
  Ouyang, \ldots, and Rybach}]{Hellsten:17}
Lars Hellsten, Brian Roark, Prasoon Goyal, Cyril Allauzen, Françoise Beaufays,
  Tom Ouyang, \ldots, and David Rybach. 2017.
\newblock Transliterated mobile keyboard input via weighted finite-state
  transducers.
\newblock In \emph{Proceedings of the 13th International Conference on Finite
  State Methods and Natural Language Processing}, pages 10--19.

\bibitem[{Henzen et~al.(1876)Henzen, Bormann, and Battista}]{CIL6}
Wilhelm Henzen, Eugen Bormann, and Giovanni~Rossi Battista, editors. 1876.
\newblock \emph{Corpus Inscriptorum Latinarum: Inscriptiones Urbis Romae
  Latinae}, volume~6.
\newblock Berolini.

\bibitem[{Hochreiter and Schmidhuber(1997)}]{Hochreiter:Schmidhuber:97}
Sepp Hochreiter and Jürgen Schmidhuber. 1997.
\newblock Long short-term memory.
\newblock \emph{Neural Computation}, 9(8):1735--1780.

\bibitem[{Jelinek(1997)}]{Jelinek:97}
Frederick Jelinek. 1997.
\newblock \emph{Statistical Methods for Speech Recognition}.
\newblock MIT Press.

\bibitem[{Kingma and Ba(2015)}]{Kingma:Ba:15}
Diederik~P. Kingma and Jimmy Ba. 2015.
\newblock Adam: a method for stochastic optimization.
\newblock In \emph{3rd International Conference on Learning Representations:
  Conference Track Proceedings}.

\bibitem[{Liang and Klein(2009)}]{Liang:Klein:09}
Percy Liang and Dan Klein. 2009.
\newblock Online {EM} for unsupervised models.
\newblock In \emph{Proceedings of Human Language Technologies: The 2009 Annual
  Conference of the North American Chapter of the Association for Computational
  Linguistics}, pages 611--619.

\bibitem[{Liu et~al.(2011)Liu, Weng, Wang, and Liu}]{Liu:11}
Fei Liu, Fuliang Weng, Bingqing Wang, and Yang Liu. 2011.
\newblock Insertion, deletion, or substitution? {N}ormalizing text messages
  without pre-categorization nor supervision.
\newblock In \emph{Proceedings of the 49th Annual Meeting of the Association
  for Computational Linguistics: Human Language Technologies}, pages 71--76.

\bibitem[{McCulloch(2019)}]{McCulloch:19}
Gretchen McCulloch. 2019.
\newblock \emph{Because Internet: Understanding the New Rules of Language}.
\newblock Riverhead Books.

\bibitem[{Merhav and Ash(2018)}]{Merhav:Ash:18}
Yuval Merhav and Stephen Ash. 2018.
\newblock Design challenges in named entity transliteration.
\newblock In \emph{Proceedings of the 27th International Conference on
  Computational Linguistics}, pages 630--640.

\bibitem[{Mohri(2009)}]{Mohri:09}
Mehryar Mohri. 2009.
\newblock Weighted automata algorithms.
\newblock In Manfred Droste, Werner Kuich, and Heiko Vogler, editors,
  \emph{Handbook of Weighted Automata}, pages 213--254. Springer.

\bibitem[{Mohri et~al.(2002)Mohri, Pereira, and Riley}]{Mohri:02}
Mehryar Mohri, Fernando Pereira, and Michael Riley. 2002.
\newblock Weighted finite-state transducers in speech recognition.
\newblock \emph{Computer Speech \& Language}, 16(1):69--88.

\bibitem[{Ney et~al.(1994)Ney, Essen, and Kneser}]{Ney:94}
Hermann Ney, Ute Essen, and Reinhard Kneser. 1994.
\newblock On structuring probabilistic dependences in stochastic language
  modelling.
\newblock \emph{Computer Speech \& Language}, 8(1):1--38.

\bibitem[{Ng et~al.(2017)Ng, Gorman, and Sproat}]{Ng:17}
Axel~H. Ng, Kyle Gorman, and Richard Sproat. 2017.
\newblock Minimally supervised written-to-spoken text normalization.
\newblock In \emph{IEEE Automatic Speech Recognition and Understanding
  Workshop}, pages 665--670.

\bibitem[{Novak et~al.(2016)Novak, Minematsu, and Hirose}]{Novak:16}
Josef~Robert Novak, Nobuaki Minematsu, and Keikichi Hirose. 2016.
\newblock Phonetisaurus: exploring grapheme-to-phoneme conversion with joint
  n-grams models in the {WFST} framework.
\newblock \emph{Natural Language Engineering}, 22(6):907--938.

\bibitem[{Pennell and Liu(2010)}]{Pennell:Liu:2010}
Deana Pennell and Yang Liu. 2010.
\newblock Normalization of text messages for text-to-speech.
\newblock In \emph{IEEE International Conference on Acoustics, Speech and
  Signal Processing}, pages 4842--4845.

\bibitem[{Ritchie et~al.(2019)Ritchie, Sproat, Gorman, van Esch, Schallhart,
  Nikos, \ldots, and Mahon}]{Ritchie:19}
Sandy Ritchie, Richard Sproat, Kyle Gorman, Daan van Esch, Christian
  Schallhart, Bampounis Nikos, \ldots, and Eoin Mahon. 2019.
\newblock Unified verbalization for speech recognition \& synthesis across
  languages.
\newblock In \emph{Proceedings of INTERSPEECH}, pages 3530--3534.

\bibitem[{Roark and Sproat(2014)}]{Roark:Sproat:14}
Brian Roark and Richard Sproat. 2014.
\newblock Hippocratic abbreviation expansion.
\newblock In \emph{Proceedings of the 52nd Annual Meeting of the Association
  for Computational Linguistics (Volume 2: Short Papers)}, pages 364--369.

\bibitem[{Roark et~al.(2012)Roark, Sproat, Allauzen, Riley, Sorensen, and
  Tai}]{Roark:12}
Brian Roark, Richard Sproat, Cyril Allauzen, Michael Riley, Jeffrey Sorensen,
  and Terry Tai. 2012.
\newblock The {OpenGrm} open-source finite-state grammar software libraries.
\newblock In \emph{Proceedings of the ACL 2012 System Demonstrations}, pages
  61--66.

\bibitem[{Sproat et~al.(2001)Sproat, Black, Chen, Kumar, Ostendorf, and
  Richards}]{Sproat:01}
Richard Sproat, Alan~W. Black, Stanley Chen, Shankar Kumar, Mari Ostendorf, and
  Christopher Richards. 2001.
\newblock Normalization of non-standard words.
\newblock \emph{Computer Speech \& Language}, 15(3):287--333.

\bibitem[{Stolcke(1998)}]{Stolcke:98}
Andreas Stolcke. 1998.
\newblock Entropy-based pruning of backoff language models.
\newblock In \emph{Proceedings of the {DARPA} {Broadcast} {News} and
  {Understanding} {Workshop}}, pages 270--274.

\bibitem[{Taylor(2009)}]{Taylor:09}
Paul Taylor. 2009.
\newblock \emph{Text-to-Speech Synthesis}.
\newblock Cambridge University Press.

\bibitem[{Yang and Eisenstein(2013)}]{Yang:Eisenstein:13}
Yi~Yang and Jacob Eisenstein. 2013.
\newblock A log-linear model for unsupervised text normalization.
\newblock In \emph{Proceedings of the 2013 Conference on Empirical Methods in
  Natural Language Processing}, pages 61--72.

\bibitem[{Zhang et~al.(2019)Zhang, Sproat, Ng, Stahlberg, Peng, Gorman, and
  Roark}]{Zhang:19}
Hao Zhang, Richard Sproat, Axel~H. Ng, Felix Stahlberg, Xiaochang Peng, Kyle
  Gorman, and Brian Roark. 2019.
\newblock Neural models of text normalization for speech applications.
\newblock \emph{Computational Linguistics}, 45(2):293--337.

\bibitem[{Żelasko(2018)}]{Zelasko:18}
Piotr Żelasko. 2018.
\newblock Expanding abbreviations in a strongly-inflected language: are
  morphosyntactic tags sufficient?
\newblock In \emph{Proceedings of the Eleventh International Conference on
  Language Resources and Evaluation}, pages 1880--1884.

\end{thebibliography}
\bibliographystyle{acl_natbib}
\end{document}